\documentclass[journal]{IEEEtran}
\usepackage{amsmath,epsfig,amssymb}
\usepackage[american]{babel}
\usepackage{booktabs}
\usepackage{hyperref}
\usepackage{enumitem}
\usepackage{array}
\usepackage{tabularx}
\usepackage{caption}
\usepackage{lipsum}
\usepackage{graphicx}
\usepackage{multirow}
\usepackage[noabbrev]{cleveref}
\usepackage[T1]{fontenc}
\usepackage{color}
\ifCLASSOPTIONcompsoc
\usepackage[caption=false, font=normalsize, labelfont=sf, textfont=sf]{subfig}
\else
\usepackage[caption=false, font=footnotesize]{subfig}
\fi

\begin{document}
%
\title{MARTA GANs: Unsupervised Representation Learning for
	Remote Sensing Image Classification}
%
%
%

\author{
	Daoyu Lin, Kun Fu, Yang Wang, Guangluan Xu, and Xian Sun 
	\thanks{This work was supported in part by the National Natural Science Foundation of China under Grant No.41501485 and No.61331017. (Corresponding author: Kun Fu.)}
	\thanks{The authors are with the Key Laboratory of Technology in Geo-spatial
	Information Processing and Application System, Chinese Academy of Sciences, Beijing, 100190, China (e-mail: lindaoyu15@mails.ucas.ac.cn; fukun@mail.ie.ac.cn; primular@163.com; gluanxu@mail.ie.ac.cn; sunxian@mail.ie.ac.cn).}

	

}



%


\maketitle
%
\begin{abstract}
With the development of deep learning, supervised learning has frequently been adopted to classify remotely sensed images using convolutional networks (CNNs). However, due to the limited amount of labeled data available, supervised learning is often difficult to carry out. Therefore, we proposed an unsupervised model called multiple-layer feature-matching generative adversarial networks (MARTA GANs) to learn a representation using only unlabeled data. MARTA GANs consists of both a generative model $G$ and a discriminative model $D$. We treat $D$ as a feature extractor. To fit the complex properties of remote sensing data, we use a fusion layer to merge the mid-level and global features. $G$ can produce numerous images that are similar to the training data; therefore, $D$ can learn better representations of remotely sensed images using the training data provided by $G$. The classification results on two widely used remote sensing image databases show that the proposed method significantly improves the classification performance compared with other state-of-the-art methods.

\end{abstract}


%
\IEEEpeerreviewmaketitle

\section{Introduction}
\label{introduction}
As satellite imaging techniques improve, an ever-growing number of high-resolution satellite images provided by special satellite sensors have become available. It is urgent to be able to interpret these massive image repositories in automatic and accurate ways. In recent decades, scene classification has become a hot topic and is now a fundamental method for land-resource management and urban planning applications. Compared with other images, remote sensing images have several special features. For example, even in the same category, the objects we are interested in usually have different sizes, colors and angles. Moreover, other materials around the target area cause high intra-class variance and low inter-class variance. Therefore, learning robust and discriminative representations from remotely sensed images is difficult.

Previously, the bag of visual words (BoVW)~\cite{sivic2003video} method was frequently adopted for remote sensing scene classification. BoVW includes the following three steps: feature detection, feature description, and codebook generation. 
To overcome the problems of the orderless bag-of-features image representation, the spatial pyramid matching (SPM) model~\cite{lazebnik2006beyond} was proposed, which works by partitioning the image into increasingly fine sub-regions and computing histograms of local features found inside each sub-region. The above-mentioned methods have comprised the state of the art for several years in the remote sensing community~\cite{lienou2010semantic}, but they are based on hand-crafted features, which are difficult, time-consuming, and require domain expertise to produce.
\begin{figure}[t]
	\begin{center}
		\includegraphics[width=0.5\textwidth]{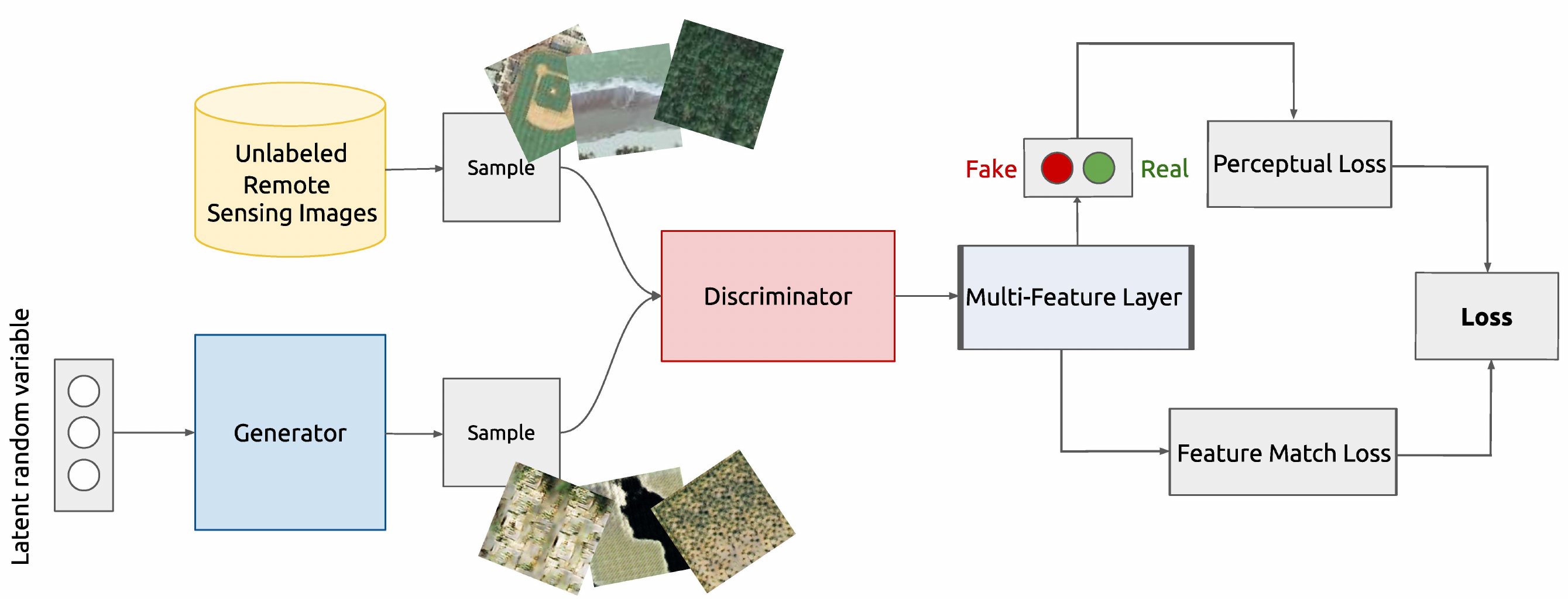}
	\end{center}
	\vspace{-1em}
	\caption{ \label{overview}Overview of the proposed approach. The discriminator~(2) learns to make classifications between real and synthesized images, while the generator~(1) learns to fool the discriminator.
		\vspace{-3mm}
	}
\end{figure}

Deep learning algorithms
 can learn high-level semantic features automatically rather than requiring handcrafted features. Some approaches~\cite{penatti2015deep,nogueira2017towards} based on convolutional neural networks (CNNs)~\cite{krizhevsky2012imagenet} have achieved success in remote sensing scene classification, but those methods usually require an enormous amount of labeled training data or are fine-tuned from pre-trained CNNs.

Several unsupervised representation learning algorithms have been based on the autoencoder~\cite{vincent2010stacked, makhzani2013k}, which receives corrupted data as input and is trained to predict the original, uncorrupted input. Although training the autoencoder requires only unlabeled data, input reconstruction may not be the ideal metric for learning a general-purpose representation. The concept of Generative Adversarial Networks (GANs)~\cite{Goodfellow2014} is one of the most exciting unsupervised algorithm ideas to appear in recent years; its purpose is to learn a generative distribution of data through a two-player minimax game. In subsequent work, a deep convolutional GAN (DCGAN)~\cite{Radford2015} achieved a high level of performance on image synthesis tasks, showing that its latent representation space captures important variation factors.

\begin{figure*}[t]
	\centering
	\subfloat[generator] {\label{fig:gen}
		\includegraphics[width=0.4\linewidth]{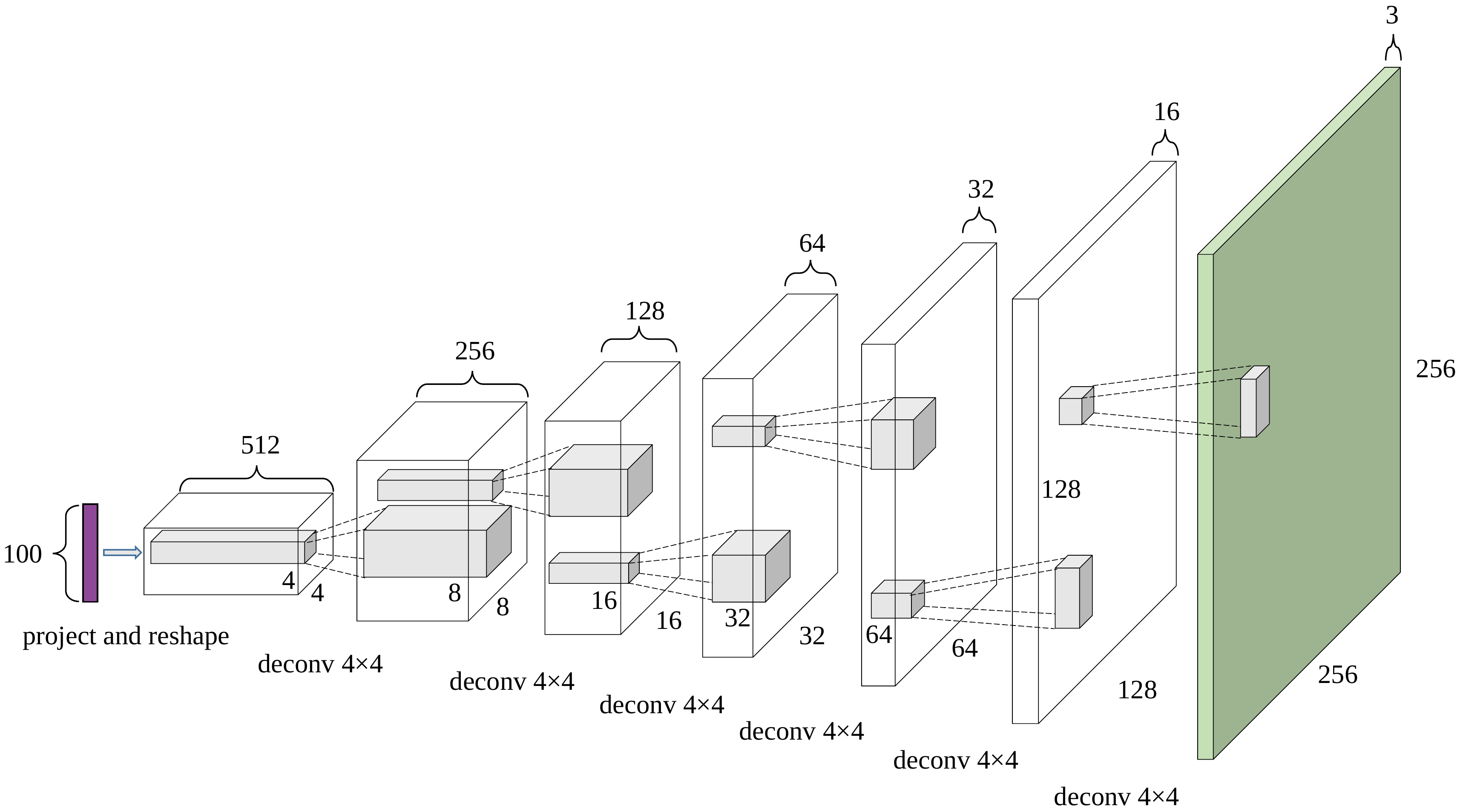}}
	\hfill
	\subfloat[discriminator]{\label{fig:disc}
		\includegraphics[width=0.58\linewidth]{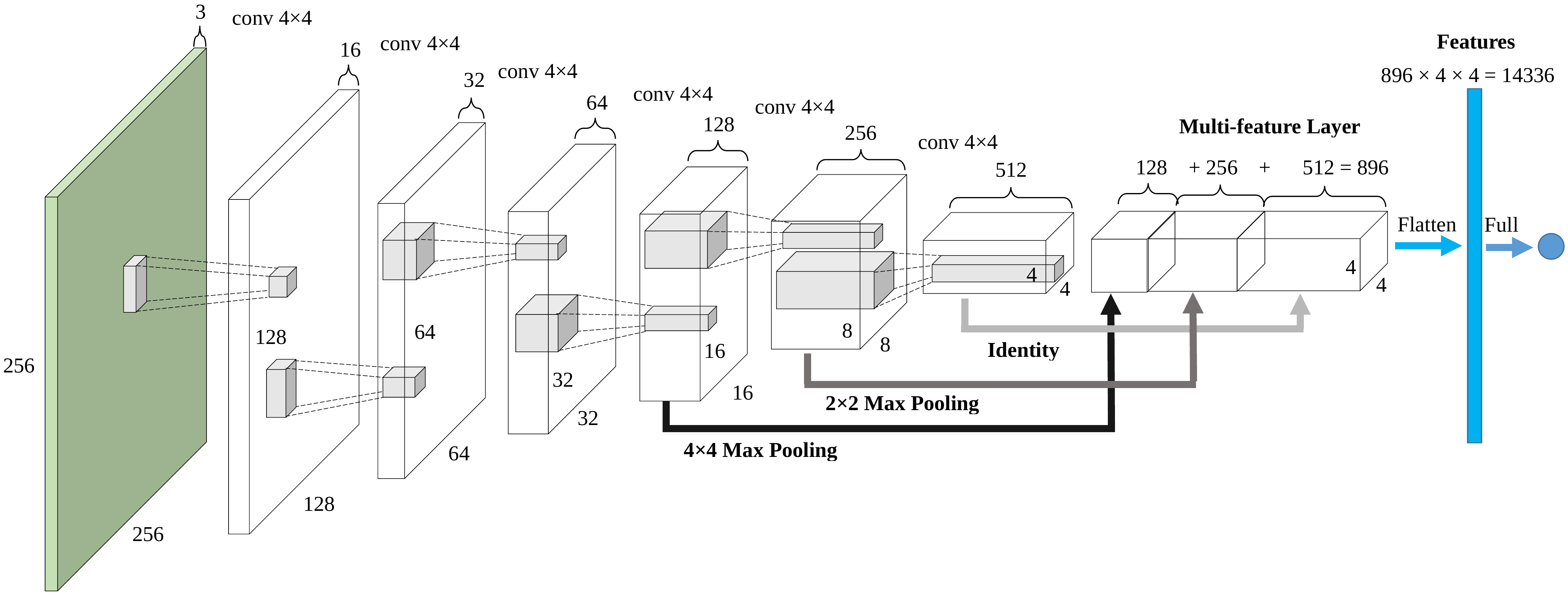}}
		
	\caption{Network architectures of a generator and a discriminator: (a) a MARTA GANs generator is used for the UC-Merced Land-use dataset. The input is a 100-dimensional uniform distribution $p_z(z)$ and the output is a $256 \times 256$-pixel RGB image; (b) a MARTA GANs discriminator is used for the UC-Merced Land-use dataset. The discriminator is treated as a feature extractor to extract features from the multi-feature layer.}
	\label{model} 
	\vspace{-4mm}
\end{figure*}

GANs is a promising unsupervised learning method, yet thus far, it has rarely been applied in the remote sensing field. Due to the tremendous volume of remote sensing images, it would be prohibitively time-consuming and expensive to label all the data. To tackle this issue, GANs would be the excellent choice because it is an unsupervised learning method in which the required quantities of training data would be provided by its generator. Therefore, in this paper, we propose a multiple-layer feature-matching generative adversarial networks 
(MARTA GANs) model to learn the representation of remote sensing images using unlabeled data.

Although based on DCGAN, our approach is rather different in the following aspects. 1) DCGAN can, at most, produce images with a $64\times64$ resolution, while our approach can produce remote sensing images with a resolution of $256\times256$ by adding two deconvolutional layers in the generator; 
2) To avoid the problem of such deconvolutional layers producing checkerboard artifacts, the kernel sizes of our networks are $4\times4$, while those in DCGAN are $5\times5$; 3) We propose a multi-feature layer to aggregate the mid- and high-level information; 4) We combine both the perceptual loss and feature matching loss to produce more accurate fake images. Based on the improvements above, our method can realize the better representation of remote sensing images among all methods. Fig.~\ref{overview} shows the overall model.


The contributions of this paper are the following:

\begin{enumerate}
	\item To our knowledge, this is the first time that GANs have been applied to classify unsupervised remote sensing images.
	\item The results of experiments on the UC-Merced Land-use and Brazilian Coffee Scenes datasets showed that the proposed algorithm outperforms state-of-the-art unsupervised algorithms in terms of overall classification accuracy.
	\item We propose a multi-feature layer by combining perceptual loss and loss of feature matching to learn better image representations.
\end{enumerate}

\section{Method}

\label{sec:method}
A GAN is most straightforward to apply when the involved models are both multilayer perceptrons; however, to apply a GAN to remote sensing images, we used CNNs for both the generator and discriminator in this work. The generator network directly produces samples $x=G(z;\theta_g)$ with parameters $\theta_g$ and $z$, where $z$ obeys a prior noise distribution $p_z(z$). Its adversary, the discriminator network, attempts to distinguish between samples drawn from the training data and samples created by the generator. The discriminator emits a probability value denoted by $D(x;\theta_d)$ with parameters $\theta_d$, indicating the probability that $x$ is a real training example rather than a fake sample drawn from the generator. During the classification task, the discriminative model $D$ is regarded as the feature extractor. Then, additional training data so that the discriminator can learn a better representation is provided by the generative model $G$.
\vspace{-1em}
\subsection{Training the discriminator}
When training the discriminator, the weights of the generator are fixed. The goals of training the discriminator $D(x)$ are as follows:

\begin{enumerate}
	\item Maximize $D(x)$ for every image from the real training examples.
	\item Minimize $D(x)$ for every image from the fake samples drawn from the generator.
\end{enumerate}

Therefore, the objective function of training discriminator is to maximize:
\begin{equation}
{\mathbb E}_{x\sim p_{\rm data}(x)} \log D(x) +
{\mathbb E}_{z\sim p_z(z)}[\log (1-D(G(z)))] \,.
\label{eq:minmax}
\end{equation}

\begin{figure*} 
	\centering
	\subfloat[real images] {\label{fig:uc_real}
		\includegraphics[width=0.48\linewidth]{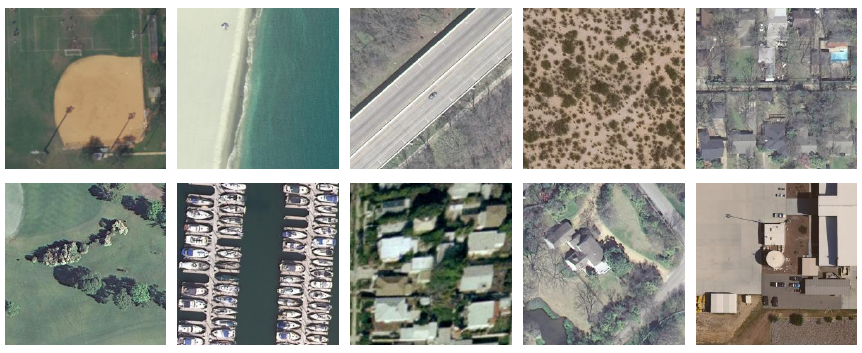}}
	\hfill
	\subfloat[fake images]{\label{fig:uc_fake}
		\includegraphics[width=0.48\linewidth]{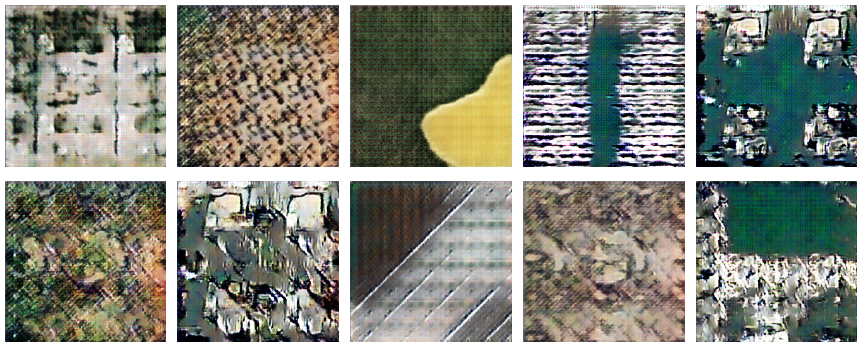}}
	
	\caption{Part of exemplary images. (a) Ten random images from UC-Merced data set. (b) Exemplary images produced by generator trained on UC-Merced using the $\ell _{final}$ (Eqn.~\ref{eq:finalloss}) objective.}
	\label{ucland} 
	\vspace{-2mm}
\end{figure*}
\vspace{-1em}
\subsection{Training the generator}
When training the generator, the weights of the discriminator are fixed. The goal of training the generator $G(z)$ is to produce samples that fool $D$. The output of the generator is an image that can be used as the input for the discriminator. Therefore, the generator wants to maximize $D(G(z))$ (or equivalently, minimize $1-D(G(z))$) because $D$ is a probability estimate that ranges only between 0 and 1. We call this concept perceptual loss; it encourages the reconstructed image to be similar to the samples drawn from the training set by minimizing the perceptual loss.
\begin{equation}
\ell_{perceptual}={\mathbb E}_{{z}\sim p_z({z})}[\log (1-D(G({z})))].
\label{eq:perceptual}
\end{equation}

In summary, the discriminator $D$ is shown an image produced from the generator $G$ and adjusts its parameters to make its output, $D(G(Z))$, larger. But $G(Z)$ will train itself to produce images that fool $D$ into thinking they are real. It does this by getting the gradient of $D$ with respect to each sample it produces. In other words, the $G$ is trying to minimize the output while $D$ is trying to maximize it; consequently, it is a minimax game that is defined as follows:
\vspace{-1.5mm}
\begin{align}
\min_G \max_D V(D,G)=&{\mathbb E}_{x\sim p_{\rm data}(x)} \log D(x) +\nonumber\\
&{\mathbb E}_{z\sim p_z(z)}[\log (1-D(G(z)))]\,.
\label{eq:minmax}
\end{align}

To make the images generated by generator more similar to the real images, we train the generator to match the expected values of the features in the multi-feature layer of the discriminator. Letting $f(x)$ denote activations on the multi-feature layer of the discriminator, the loss of feature matching for the generator is defined as follows:
\begin{equation}
\ell_{feature\_match}=||{\mathbb E}_{x\sim p_{\rm data}(x)} f(x)-{\mathbb E}_{z\sim p_z(z)}f(G(z))||_2^2 \,.
\label{eq:featurematchloss}
\end{equation}
Therefore, our final object (the combination of Eqn. \ref{eq:perceptual} and Eqn. \ref{eq:featurematchloss}) for training the generator is to minimize Eqn. \ref{eq:finalloss}~. 
\begin{equation}
\ell_{final}=\ell_{perceptual}+ \ell_{feature\_matching}
\label{eq:finalloss}.
\end{equation}

\subsection{Network architectures}
\label{subsect:net arc}


The details of the generator and discriminator in MARTA GANs are as follows:

The generator takes 100 random numbers drawn from a uniform distribution as input.
Then, the result is reshaped into a four-dimensional tensor. We used six deconvolutional layers in our generator to learn its own spatial upsampling and upsample the $4\times 4$ feature maps to $256\times256$ remote sensing images.
Fig.~\ref{fig:gen} shows a visualization of the generator.

%


For the discriminator, the first layer takes input images, including both real and synthesized images. We use convolutions in our discriminator which allows it to learn its own spatial downsampling. As shown in Fig.~\ref{fig:disc}, by performing $4\times4$ max pooling, $2\times2$ max pooling and the identity function separately in the last three convolutional layers, we can produce feature maps that have the same spatial size, $4\times4$. Then, we concatenate the  $4\times4$ feature maps through channel dimension in the multi-feature layer.
Finally, the  multi-feature layer is flattened and fed into a single sigmoid output.
The multi-feature layer includes two functions: 1) the features used for classification are extracted from the flatted multi-feature layer; 2) when training the generator, we use feature matching loss (Eqn.~\ref{eq:featurematchloss}) to evaluate the similarities of the features between the fake and real images in the flatted multi-feature layer. 

We set the kernel sizes to $4\times4$ and the stride to 2 in all the convolutional and deconvolutional layers, because the deconvolutional layers can avoid uneven overlap when the kernel size is divisible by the stride~\cite{odena2016deconvolution}. In the generator, all layers use ReLU activation except for the output layer, which uses the tanh function. We use LeakyReLU activation in the discriminator for all the convolutional layers; the slope of the leak was set to 0.2.
We used batch normalization in both the generator and the discriminator, and the decay factor was 0.9.


\section{Experiments}
\label{subsect:experiments}
To verify the effectiveness of the proposed method, we trained MARTA GANs on two datasets: the UC Merced Land Use dataset~\cite{yang2010bag} and the Brazilian Coffee Scenes dataset~\cite{penatti2015deep}. We carried out experiments on both datasets using a 5-fold cross-validation protocol and a regularized linear L2-SVM as a classifier. We implemented MARTA GANs in TensorLayer~\footnote{\url{http://tensorlayer.readthedocs.io/en/latest/}}, a deep learning and reinforcement learning library extended from Google TensorFlow \cite{abadi2016tensorflow}. We scaled the input image to the range of [-1, 1] before training. All the models were trained by SGD with a batch size of 64, and we used the Adam optimizer with a learning rate of 0.0002 and a momentum term $\beta_1$ of 0.5.


\begin{table*}
	\centering
	\scriptsize 
	\caption{ Classification accuracy (\%) in the form of the means $\pm$ standard deviation bars of DCGAN and MARTA GAN for every class.. The class labels are as follows: 1 = Mobile home park, 2 = Beach, 3 = Tennis courts, 4 = Airplane, 5 = Dense residential, 6 = Harbor, 7 = Buildings, 8= Forest, 9 = Intersection, 10 = River, 11 = Sparse residential, 12 = Runway, 13 = Parking lot, 14 = Baseball diamond, 15 = Agricultural, 16 = Storage tanks, 17 = Chaparral, 18 = Golf course, 19 = Freeway, 20 = Medium residential, and 21 = Overpass.}
	\vspace{-0.5em}
	\label{tab:compare_acc}
	\begin{tabular}{|c|ccccccccccc|}
     \hline
		
		Class&1& 2 &3& 4 & 5 & 	6& 7 &8 & 9 &10&11\\ 
		\hline
		DCGAN &$85\pm5.0$ & $94\pm2.2$ & $89\pm4.2$ & $95\pm3.5$ & $82\pm2.7$ & 
			$91\pm2.2$& $78\pm2.7$ & $83\pm2.7$ & $88\pm2.7$ & $90\pm0.0$ & $79\pm2.2$ \\ 
		
		MARTA GAN &$95\pm3.5$& $100\pm0.0$ & $96\pm4.2$ & $100\pm0.0$ &$89\pm4.2$&
			$99\pm2.2$& $86\pm6.5$ & $97\pm2.7$ & $98\pm2.7$ &$94\pm2.2$& $89\pm2.2$
			\\ 
	\hline
		Class&12 & 13 & 14 & 15 & 16 &17 &18 &19 &20 &21 &\\ 
	\hline	
		DCGAN& $89\pm4.2$ & $88\pm2.7$ & $95\pm3.5$ & $78\pm4.5$ & 
		$93\pm2.7$& $88\pm2.7$ & $97\pm2.7$ & $77\pm2.7$ & $95\pm5.0$ & 	$89\pm4.2$&\\ 
		MARTA GAN&  $94\pm4.2$ & $98\pm2.7$ & $100\pm0.0$ &$85\pm5.0$
		&	$100\pm0.0$& $93\pm2.7$ & $100\pm0.0$ & $87\pm5.7$ &$97\pm5.5$&$95\pm5.0$ &

	\\ 
	\hline
	\end{tabular}
\vspace{-6mm}
\end{table*}

%
%
%
%
%
%
%
%
%
%

\subsection{UC Merced dataset}

This dataset consists of images of 21 land-use classes (100 $256\times256$-pixel images for each class).
Some of the images from this dataset are shown in Fig.~\ref{fig:uc_real}. We used a moderate data augmentation in this dataset via flipping images horizontally and vertically and rotating them by 90 degrees to increase the effective training set size. Training takes approximately 4 hours on a single NVIDIA GTX 1080 GPU.


\begin{figure}[t]
	\centering
	\subfloat[] {
		\includegraphics[width=0.24\linewidth]{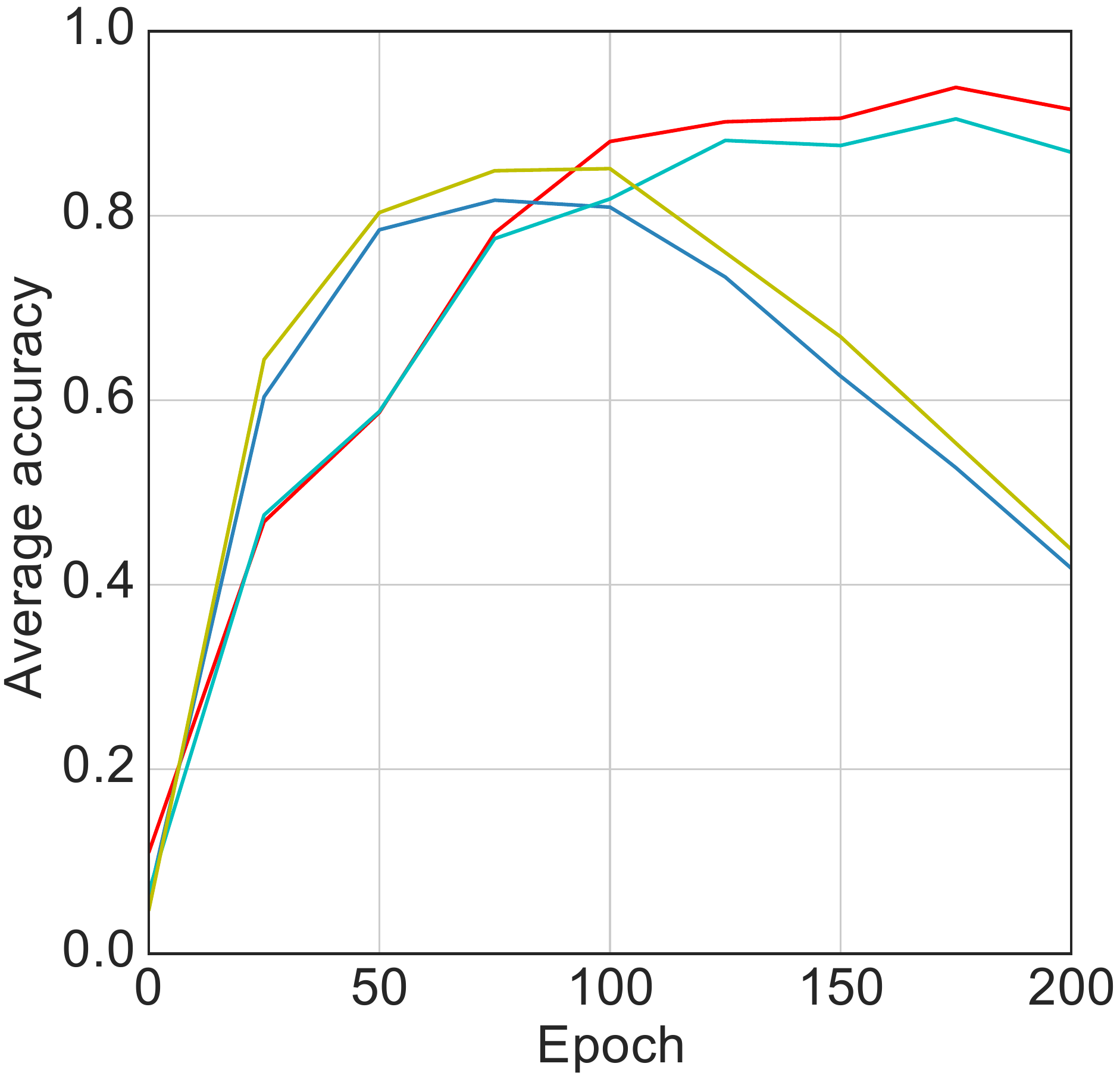}}
	\subfloat[ ]{
		\includegraphics[width=0.24\linewidth]{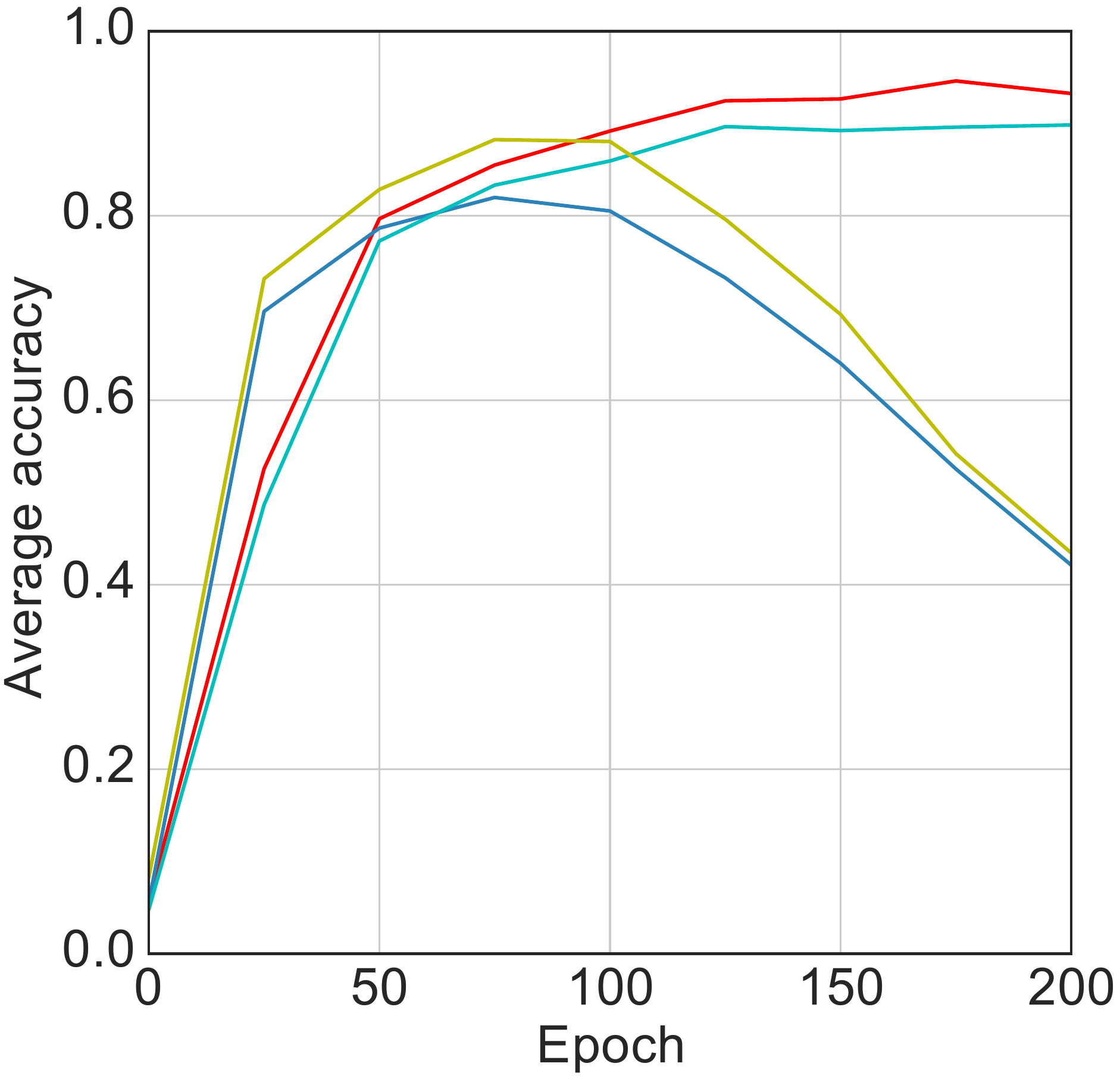}}
	\subfloat[ ]{
		\includegraphics[width=0.24\linewidth]{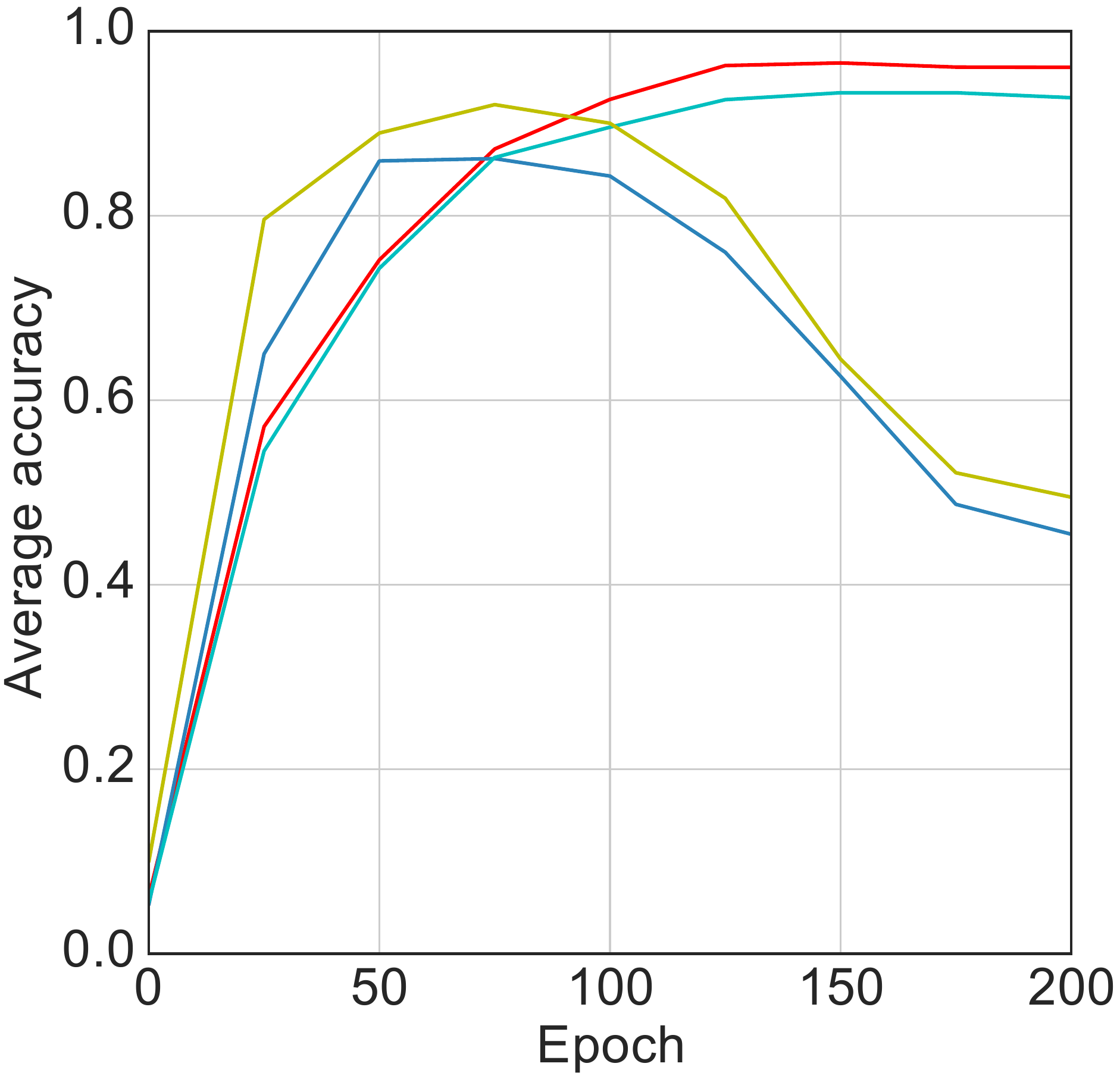}}
	\subfloat[ ]{
		\includegraphics[width=0.24\linewidth]{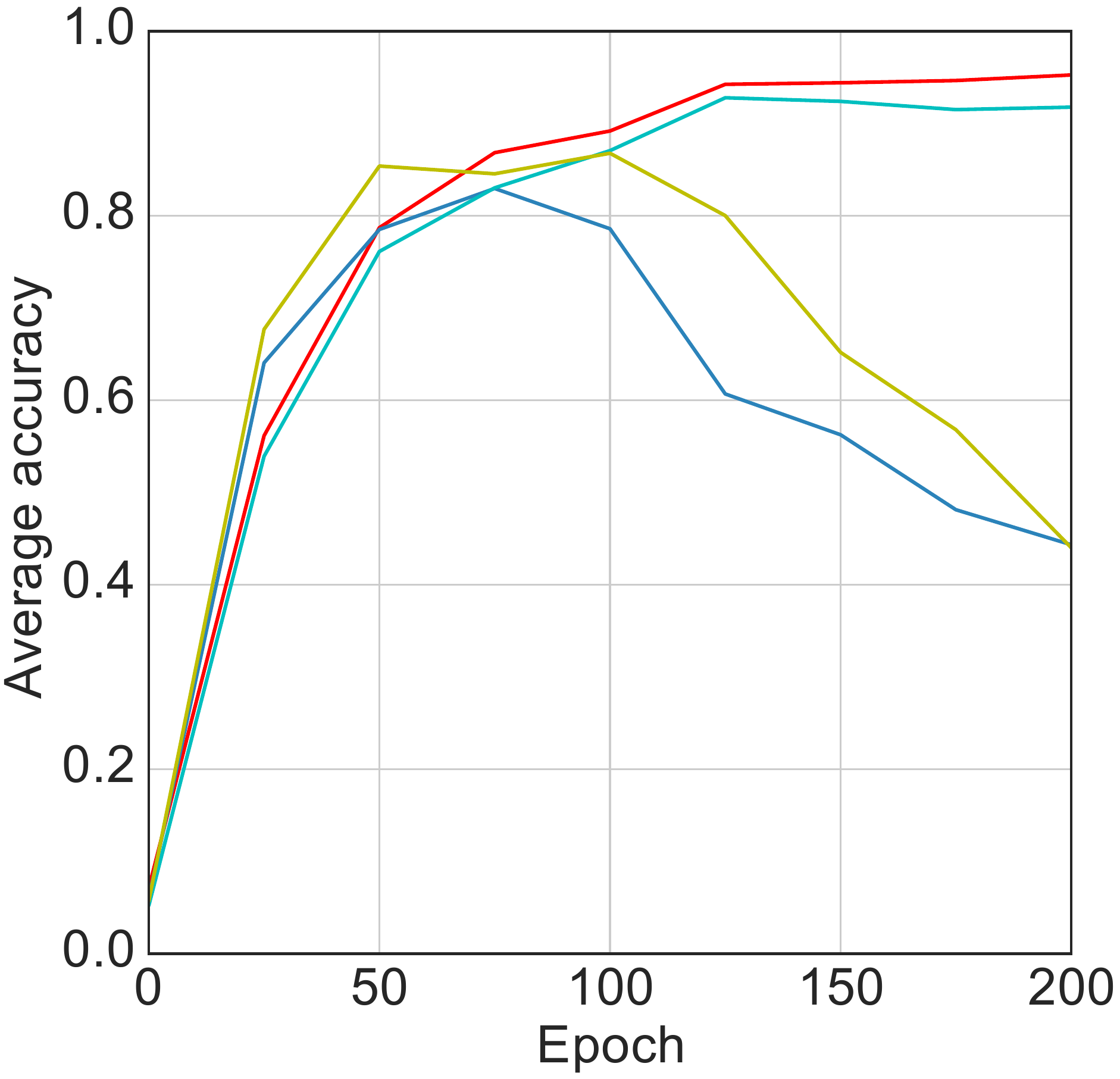}}
	\\
	\vspace{-2mm}
	\caption{The performance comparison uses different features. (a) $f_1$; (b) $f_2$; (c) $f_3$;
		(d) $f_4$. 
		The red curves: training with $\ell_{final}$ and with data augmentation;
		Cyan curves: training with $\ell_{perceptual}$ and with data augmentation;
		Yellow curves: training with $\ell_{final}$ and without data augmentation;
		Green curves: training with $\ell_{perceptual}$ and without data augmentation. }
	\vspace{-2mm}
	\label{fig:compare_acc}
\end{figure}

To evaluate the quality of the representations learned by the multi-feature layer, we trained on the UC-Merced data and extracted the features from different multi-feature layers.
To improve the clarity of the expression, we use $f_1$ to denote the features from the last convolutional layer, $f_2$ to denote features combined from the last two convolutional layers' features, and so on. Based on the results shown in Fig.~\ref{fig:compare_acc}, we found that $f_3$ achieved the highest accuracy. These results can be explained by two reasons. First, $f_3$ has the same high-level information as $f_1$ and $f_2$, but it has more mid-level information compared with $f_1$ and $f_2$. However, $f_4$ has too much low-level information, which leads to the "curse of dimensionality." Therefore, the features extracted from the last three convolutional layers in the discriminator resulted in the highest accuracy. As shown in Fig.~\ref{fig:compare_acc}, data augmentation is an effective way to reduce overfitting when training a large deep network. Augmentation generates more training image samples by rotating and flipping patches from original images. We also evaluated the performance between two types of loss: $\ell_{perceptual}$ (Eqn. \ref{eq:perceptual}) and $\ell_{final}$ (Eqn.~\ref{eq:finalloss}) and found that using $\ell_{final}$ achieved the best performance. Synthesized remote sensing images when using $\ell_{final}$ are shown in Fig.~\ref{fig:uc_fake}.

\begin{figure}[t]
	\centering
	\subfloat[] {\label{fig:real}
		\includegraphics[width=0.48\linewidth]{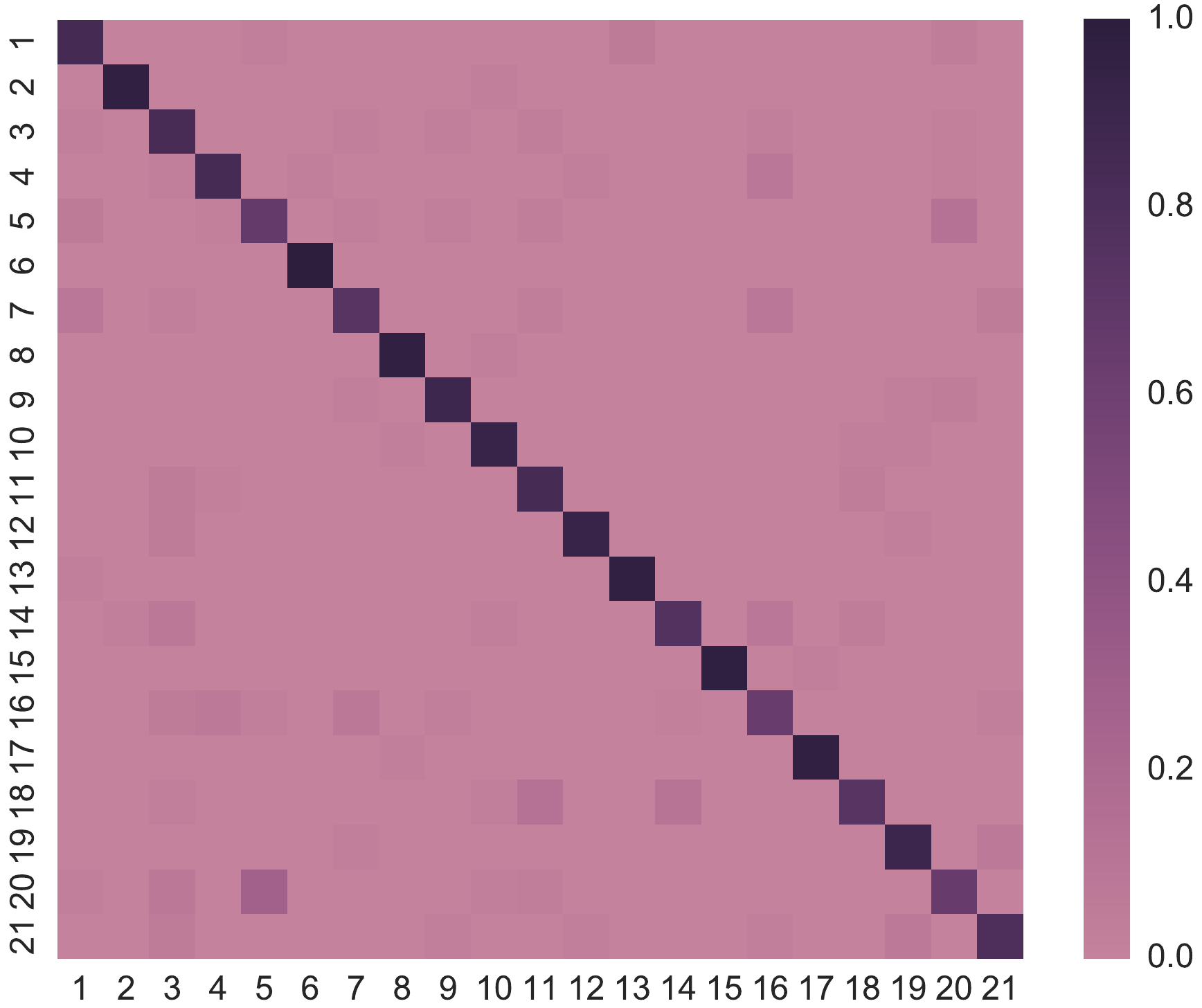}}
	\hfill
	\subfloat[]{\label{fig:fake}
		\includegraphics[width=0.48\linewidth]{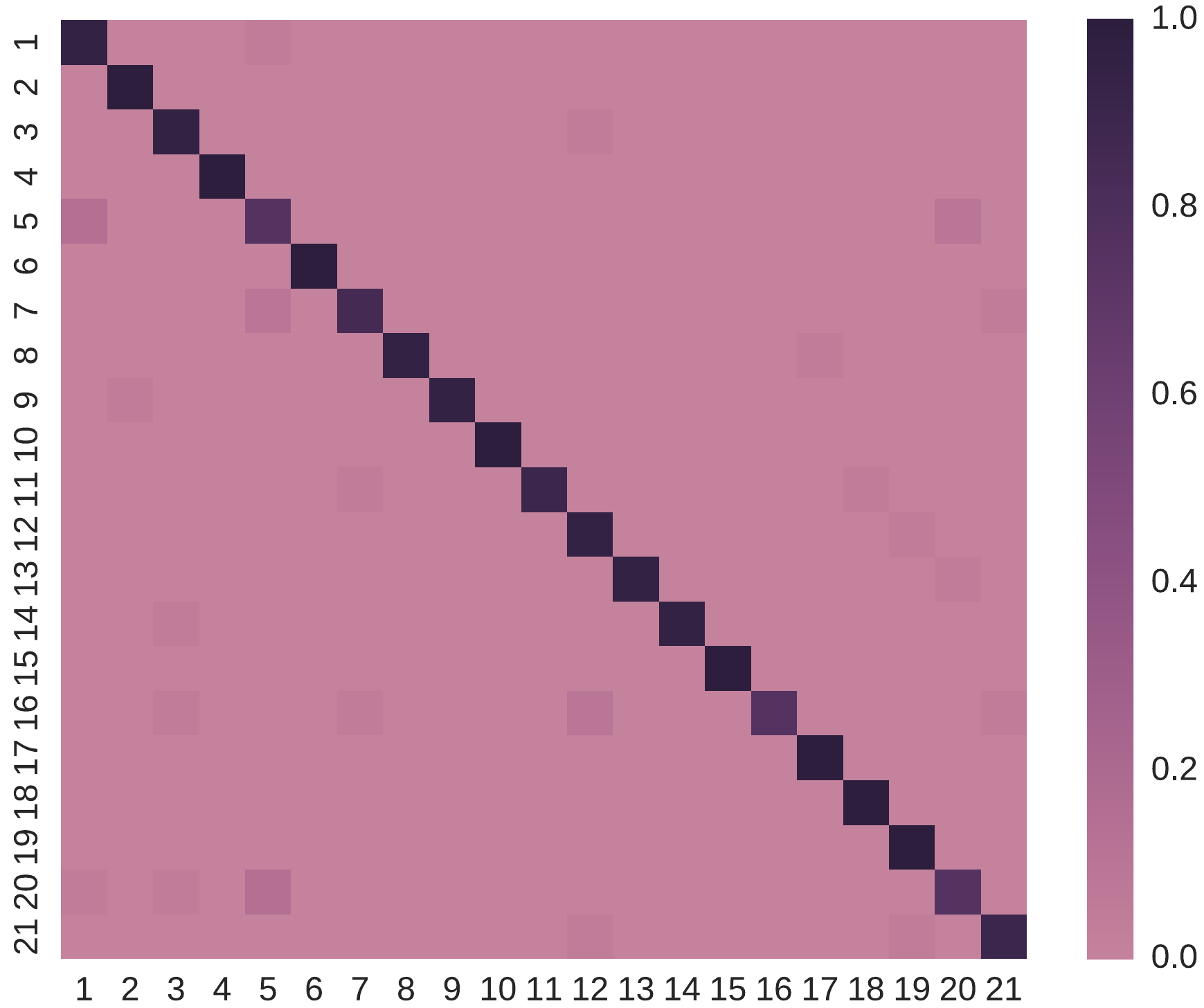}}
	\\
	\caption{Confusion matrix of (a):DCGAN, (b):MARTA GAN. The class labels are same as Table~\ref{tab:compare_acc}.}
	\label{fig:conf_matix} 
		\vspace{-5mm}
\end{figure}

Fig.~\ref{fig:conf_matix} depicts the confusion matrix of classification results for the two GAN architectures, DCGAN and MARTA GAN. DCGAN and MARTA GAN reached an overall accuracy of $87.76\pm0.64$\% and $94.86\pm0.80$\%, respectively. MARTA GAN is approximately 7\% better because it used the multi-feature layer to merge the mid-level and global features. To improve the comparison, the accuracy classification performances of the methods for each class are shown in Table~\ref{tab:compare_acc}. Compared to DCGAN, MARTA GAN achieves 100.00\% accuracy in some scene categories (e.g., Beach, Airplane, etc.). Moreover, MARTA GAN also achieves higher accuracy in some very close classes, such as dense residential, building, medium residential, sparse residential.

\begin{figure}[t]
	\centering
	\includegraphics[width=0.49\textwidth]{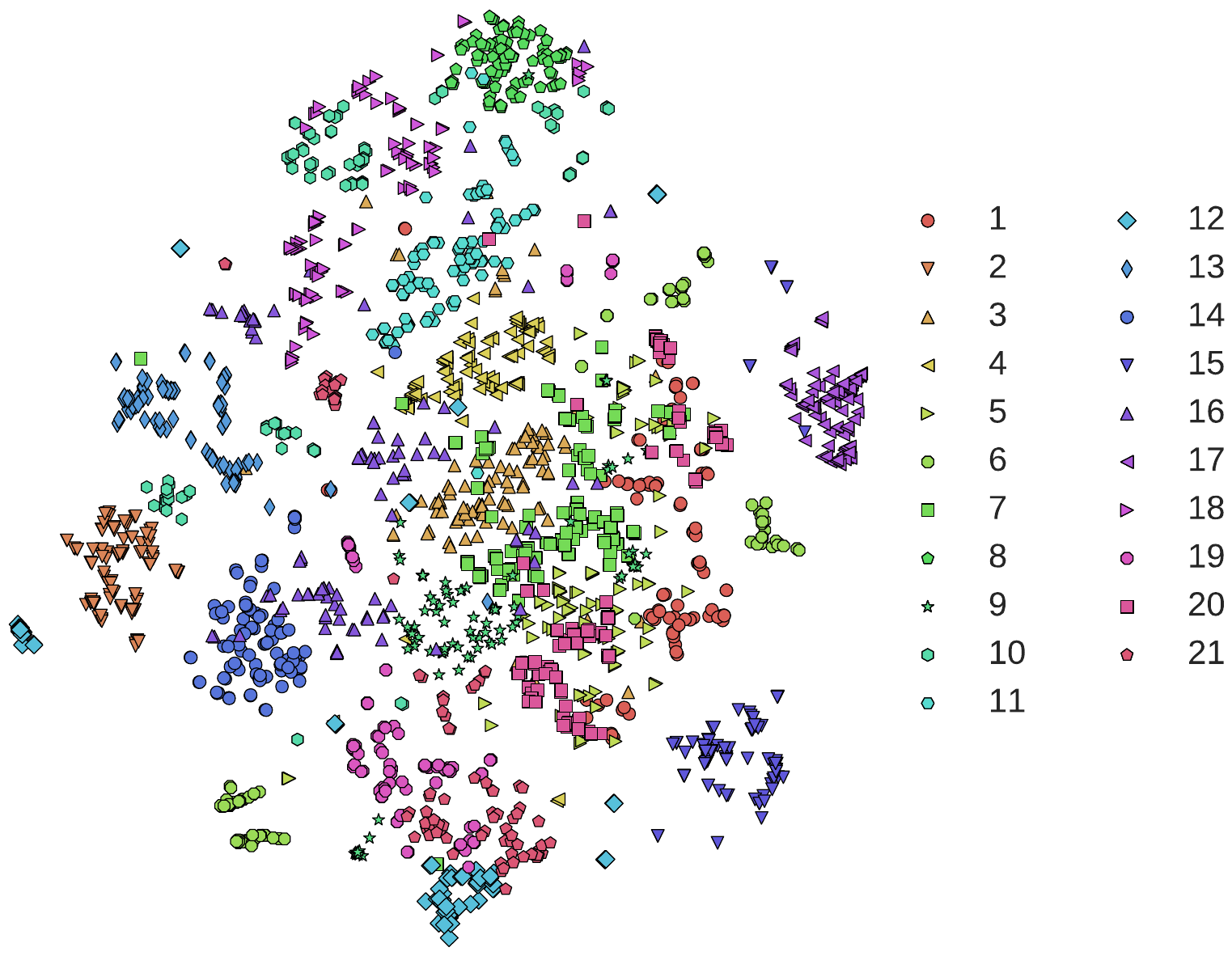}
	\vspace{-1.5em}
	\caption{2-D feature visualization of image global representations of the UC-Merced dataset.  The class labels are same as Table~\ref{tab:compare_acc}.}
	\vspace{-4mm}
	\label{fig:t_sne}
\end{figure}

\begin{figure*} 
	\centering
	\subfloat[real images] {\label{fig:coffee_real}
		\includegraphics[width=0.48\linewidth]{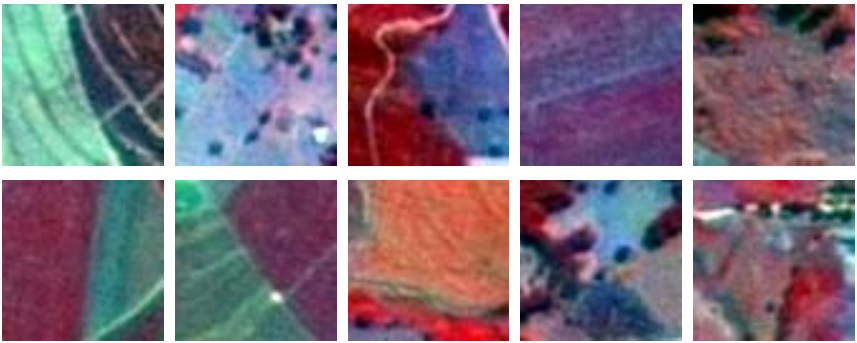}}
	\hfill
	\subfloat[fake images]{\label{fig:coffe_fake}
		\includegraphics[width=0.48\linewidth]{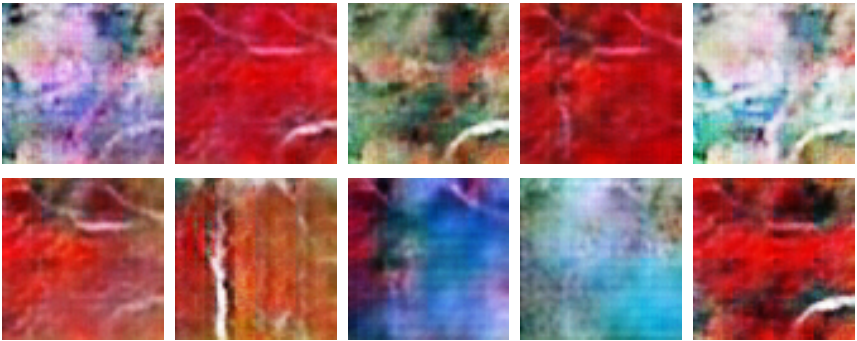}}
	\\
	\caption{Parts of exemplary images: (a) ten random images from the Brazilian Coffee Scenes dataset; (b) exemplary images produced by a generator trained on the Brazilian Coffee Scenes dataset using the $\ell _{final}$ (Eqn. \ref{eq:finalloss}) objective.}
	\label{coffee}
		\vspace{-2mm} 
\end{figure*}

In addition, we visualized the global image representations encoded via MARTA GANs features of the UC-Merced dataset. We computed the features for all the scenes of the dataset and then used the t-SNE algorithm to embed the high-dimensional features in 2-D space. The final results are shown in Fig. \ref{fig:t_sne}. This visualization shows that features extracted from the multi-feature layer contain abstract semantic information because those close classes are also very close in 2-D space.


Compared with the results of other tested methods, the method proposed in this work achieves the highest classification accuracy among the unsupervised methods. As shown in Table~\ref{tab:uc_ref}, our method outperforms the SCMF~\cite{sheng2012high} (a sparse coding based multiple-feature fusion method) by 3.82\%. When the classification accuracy of our method is compared with  LRFF~\cite{hu2015unsupervised} (an improved unsupervised feature learning algorithm based on spectral clustering), our method  outperforms LRFF by more than 4\%. While some of the supervised methods~\cite{penatti2015deep, nogueira2017towards} achieved an accuracy above 99\%, these methods are fine-tuned from pre-trained models, which are usually trained with a large amount of labeled data (such as ImageNet). Compared with those methods, our unsupervised method requires fewer parameters.



%

\begin{table}[t]
	\centering
	\scriptsize
	\caption{Overall classification accuracy (\%) of reference and proposed methods on the UC-Merced dataset and Coffee Scenes dataset. Our result is in bold.}
	\label{tab:uc_ref}
	
	\begin{tabular}{|p{1.2cm}<{\centering} | p{2.25cm} p{1.2cm} p{0.7cm}<{\centering}  p{1.4cm}<{\centering}|}
		\hline
		 DataSet&Method & Description&Parameters &Accuracy \\
		\hline
		 \multirow{5}{*}{UC-Merced}&SCMF~\cite{sheng2012high}& Unsupervised&- & $91.03\pm0.48$ \\
		&UFL-SC~\cite{hu2015unsupervised}&Unsupervised&-& $90.26\pm1.51$  \\
		&OverFeat$_L$ + Caffe~\cite{penatti2015deep}&Supervised&205M&$99.43\pm0.27$\\
		&GoogLeNet~\cite{nogueira2017towards}&Supervised&5M&$99.47\pm0.50$\\
		&\textbf{MARTA GANs}&\textbf{Unsupervised}&\textbf{2.8M}& \textbf{$94.86\pm0.80$}\\
		\hline
		 \multirow{4}{*}{Coffee}	&BIC~\cite{penatti2015deep}& Unsupervised&- & $87.03\pm1.07$ \\
	&OverFeat$_L$+OverFeat$_S$~\cite{penatti2015deep}&Supervised&289M& $83.04\pm2.00$  \\
		&CaffeNet~\cite{nogueira2017towards}&Supervised&60M& $94.45\pm1.20$  \\
		&\textbf{MARTA GANs}&\textbf{Unsupervised}&\textbf{0.18M}& \textbf{$89.86\pm0.98$}\\
		\hline
	\end{tabular}
\vspace{-1em}
\end{table}

\subsection{Brazilian Coffee Scenes dataset}
\label{subsect:coffee}
To evaluate the generalization power of our model, we also performed experiments using the Brazilian Coffee Scenes dataset~\cite{penatti2015deep}, which is a composition of scenes taken by the SPOT sensor in the green, red, and near-infrared bands. This dataset has 2,876 multispectral high-resolution scenes. It includes 1,438 tiles of coffee and 1,438 tiles of non-coffee with a $64\times64$-pixel resolution. Fig.~\ref{fig:coffee_real} shows some examples of this dataset. We did not use data augmentation on this dataset because it contains sufficient data to train the network.

%
%

%

Table~\ref{tab:uc_ref} shows the results obtained with the proposed method. In general, the results are significantly worse than those on the UC-Merced dataset, despite reducing the classification from a 21-class to a 2-class problem. Brazilian Coffee Scenes is a challenging dataset because of the high intra-class variability caused by different crop management techniques, different plant ages and spectral distortions and shadows. Nevertheless, our results are better than that of BIC~\cite{penatti2015deep}.


\vspace{-2mm}
\section{CONCLUSION}
\label{sec:discussion}
\vspace{-2mm}
This paper introduced a representation learning algorithm called MARTA GANs. In contrast to previous approaches that require supervision, MARTA GANs is completely unsupervised; it can learn interpretable representations even from challenging remote sensing datasets. In addition, MARTA GANs introduces a new multiple-feature-matching layer that learns multi-scale spatial information for high-resolution remote sensing. Other possible future extensions to the work described in this paper include: producing high-quality samples of remote sensing images using the generator and classifying remote sensing images in a semi-supervised manner to improve classification accuracy.



\bibliographystyle{IEEEtran}
\bibliography{paper}

\end{document}